\documentstyle{article}

\newenvironment{reflist}{\begin{list}{}%
{\setlength{\labelwidth}{0pt}
 \setlength{\labelsep}{0pt}
 \setlength{\leftmargin}{3em}
 \setlength{\itemindent}{-3em}
 \setlength{\listparindent}{-3em}
 \setlength{\itemsep}{6pt}}\item}{\end{list}}

\begin{document}
\begin{titlepage}
\vspace*{1.65in}
\begin{center}
{\large					Research Report AI-1994-01} \\

\vspace{.1in}

{\bf \large			An Empirically Motivated Reinterpretation \\
					    of Dependency Grammar} \\
\vspace{.1in}
{\large					    Michael A. Covington}\\
\vspace{.1in}

				       Artificial Intelligence Programs\\
				           The University of Georgia\\
				          Athens, Georgia 30602 U.S.A.\\
\end{center}
\end{titlepage}

\begin{center}
{\large \bf	       An Empirically Motivated Reinterpretation of Dependency
Grammar}\\
\vspace{.3in}
{\bf 						Michael A.\ Covington}\\
\vspace{.25in}
        				   Artificial Intelligence Programs \\

        					The University of Georgia \\
        				       Athens, Georgia 30602--7415 \\
        		 			  mcovingt@ai.uga.edu\\

\end{center}

\newcommand{\barr}[1]{$\overline{\rm #1}$}
\newcommand{\barrr}[1]{$\overline{\overline{\rm #1}}$}
\newcommand{\barrrr}[1]{$\overline{\overline{\overline{\rm #1}}}$}

\begin{abstract}
Dependency grammar is usually interpreted as equivalent to a strict form
of X--bar theory that forbids the stacking of nodes of the same bar
level (e.g., \barr{N} immediately dominating \barr{N} with the same
head).  But adequate accounts of {\em one}--anaphora and of the
semantics of multiple modifiers require such stacking and accordingly
argue against dependency grammar. Dependency grammar can be salvaged by
reinterpreting its claims about phrase structure, so that modifiers map
onto binary--branching X--bar trees rather than ``flat'' ones.
\end{abstract}

\section{Introduction}

Arguments for stacked X--bar structures (such as \barr{N} immediately
dominating \barr{N} with the same head)
are arguments against dependency grammar as normally understood.
This paper reviews the dependency grammar formalism, presents evidence
that stacked \barr{N} structures are required, and then proposes a
reinterpretation of dependency grammar to make it compatible with the
evidence.

\section{Dependency grammar}

\subsection{The formalism}

Dependency grammar (DG) describes syntactic structure in terms of links
between individual words rather than constituency trees. DG has its
roots in Arabic and Latin traditional grammar; its modern advocates
include Tesni\`{e}re (1959), Robinson (1970), Starosta (1988),
Mel'\v{c}uk (1987), Hudson (1980a, 1980b, 1990), and myself (Covington
1990).

The fundamental relation in DG is between head and dependent.  One word
(usually the main verb) is the head of the whole sentence; every other
word depends on some head, and may itself be the head of any number of
dependents.  The rules of grammar then specify what heads can take what
dependents (for example, adjectives depend on nouns, not on verbs).
Practical DGs distinguish various types of dependents (complement,
adjunct, determiner, etc.), but the details are not important for my
argument.

Figure 1 shows, in the usual notation, a dependency analysis of {\em The
old dog chased the cat into the garden.}  Here {\em chased} is the head
of the sentence;%
\footnote{But it would be completely compatible with the formalism to
postulate that the head of the sentence is a potentially empty
INFL or the like.  Then, in Fig.\ 2, the VP would be a constituent.}
{\em dog} and {\em cat} depend on {\em chased}; {\em the} and {\em old}
depend on {\em dog;} and so on.

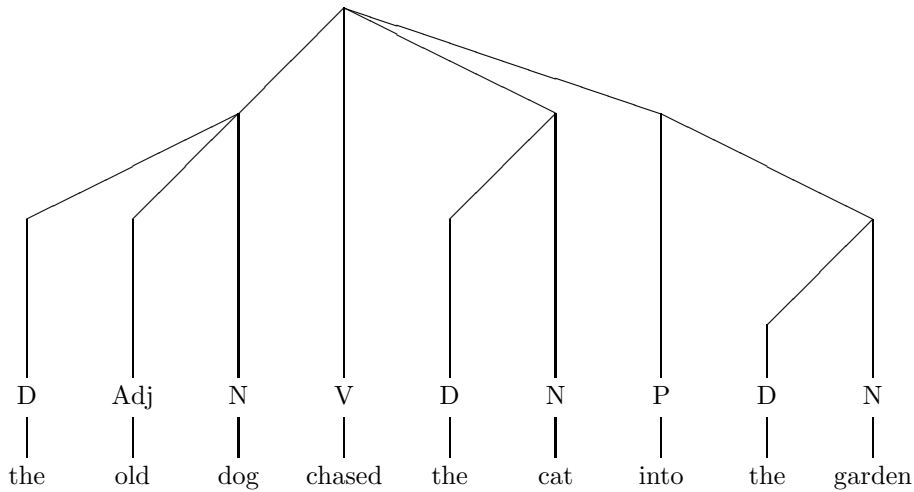
\begin{figure}
\begin{picture}(345,190)
\put(245,150){\line(2,-1){80}}
\put(325,50){\line(0,1){60}}
\put(325,40){\makebox(0,0)[b]{N}}
\put(325,35){\line(0,-1){15}}
\put(325,10){\makebox(0,0)[b]{garden}}
\put(325,110){\line(-1,-1){40}}
\put(285,50){\line(0,1){20}}
\put(285,40){\makebox(0,0)[b]{D}}
\put(285,35){\line(0,-1){15}}
\put(285,10){\makebox(0,0)[b]{the}}
\put(125,190){\line(3,-1){120}}
\put(245,50){\line(0,1){100}}
\put(245,40){\makebox(0,0)[b]{P}}
\put(245,35){\line(0,-1){15}}
\put(245,10){\makebox(0,0)[b]{into}}
\put(125,190){\line(2,-1){80}}
\put(205,50){\line(0,1){100}}
\put(205,40){\makebox(0,0)[b]{N}}
\put(205,35){\line(0,-1){15}}
\put(205,10){\makebox(0,0)[b]{cat}}
\put(205,150){\line(-1,-1){40}}
\put(165,50){\line(0,1){60}}
\put(165,40){\makebox(0,0)[b]{D}}
\put(165,35){\line(0,-1){15}}
\put(165,10){\makebox(0,0)[b]{the}}
\put(125,50){\line(0,1){140}}
\put(125,40){\makebox(0,0)[b]{V}}
\put(125,35){\line(0,-1){15}}
\put(125,10){\makebox(0,0)[b]{chased}}
\put(125,190){\line(-1,-1){40}}
\put(85,50){\line(0,1){100}}
\put(85,40){\makebox(0,0)[b]{N}}
\put(85,35){\line(0,-1){15}}
\put(85,10){\makebox(0,0)[b]{dog}}
\put(85,150){\line(-1,-1){40}}
\put(45,50){\line(0,1){60}}
\put(45,40){\makebox(0,0)[b]{Adj}}
\put(45,35){\line(0,-1){15}}
\put(45,10){\makebox(0,0)[b]{old}}
\put(85,150){\line(-2,-1){80}}
\put(5,50){\line(0,1){60}}
\put(5,40){\makebox(0,0)[b]{D}}
\put(5,35){\line(0,-1){15}}
\put(5,10){\makebox(0,0)[b]{the}}
\end{picture}
\caption{A dependency analysis.  Heads are connected to dependents by
downward--sloping lines.}
\end{figure}

\subsection{Constituency in DG}

Dependency grammar still recognizes constituents, but they are a defined
rather than a basic concept.  The usual definition is that a
constituent consists of any word plus
all its dependents, their dependents, and so on recursively.
(Tesni\`{e}re calls such a constituent a {\sc n\oe{}ud}.)
Thus the constituents in Figure 1
are (in addition to the individual words):
\begin{flushleft}
{\em the old dog} (headed by {\em dog})\\
{\em the garden} (headed by {\em garden})\\
{\em into the garden} (headed by {\em into})\\
{\em the old dog chased the cat into the garden} (headed by {\em chased}).
\end{flushleft}
There is a rule that, at least in English, every constituent must
be a contiguous string of words (Robinson 1970; Hudson 1990:114--120).

Because of its assertion that every constituent has a head, DG formalism
is equivalent to a particular strict form of X--bar theory in which:
\begin{itemize}
\item There is only one non--terminal bar level (i.e., X and \barr{X},
but not \barrr{X}, \barrrr{X}, etc.);

\item Apart from bar level, X and the \barr{X} immediately dominating it
cannot differ in any way, because they are ``really'' the same node;

\item There is no ``stacking'' of \barr{X} nodes (an \barr{X} node cannot
dominate another \barr{X} with the same head).
\end{itemize}
The third of these observations is the critical one: structures of the
form
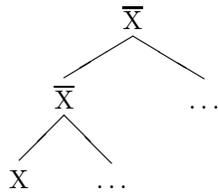
\begin{flushleft}
\begin{picture}(125,97)
\put(78,77.5){\makebox(0,0)[b]{\barr{X}}}
\put(78,75){\line(5,-3){27}}
\put(78,75){\line(-5,-3){26}}
\put(105,47.5){\makebox(0,0)[b]{$\ldots$}}
\put(52,47.5){\makebox(0,0)[b]{\barr{X}}}
\put(52,45){\line(1,-1){18}}
\put(52,45){\line(-1,-1){17}}
\put(70,17.5){\makebox(0,0)[b]{$\ldots$}}
\put(35,17.5){\makebox(0,0)[b]{X}}
\end{picture}
\end{flushleft}
are ruled out.
Figure 2 shows Figure 1 recast into X--bar theory according to this
interpretation.

\begin{figure} 
  \begin{picture}(350.0,174.0)(0,0)

\put(162.847,161.0){\makebox(1,1){\rule[-1ex]{0ex}{3ex}{\normalsize\barr{V}}}}
   \put(162.847,156.0){\line(-6,-1){105.0}}
   \put(162.847,156.0){\line(-2,-1){31.597}}
   \put(162.847,156.0){\line(2,-1){31.597}}
   \put(162.847,156.0){\line(6,-1){105.0}}

\put(48.611,131.0){\makebox(1,1){\rule[-1ex]{0ex}{3ex}{\normalsize\barr{N}}}}
   \put(48.611,126.0){\line(-5,-2){38.889}}
   \put(48.611,126.0){\line(0,-1){16.0}}
   \put(48.611,126.0){\line(5,-2){38.889}}
   \put(9.722,101.0){\makebox(1,1){\rule[-1ex]{0ex}{3ex}{\normalsize
\barr{D}}}}
   \put(9.722,96.0){\line(0,-1){16.0}}
   \put(9.722,71.0){\makebox(1,1){\rule[-1ex]{0ex}{3ex}{\normalsize D}}}
   \put(9.722,66.0){\line(0,-1){46.0}}
   \put(9.722,11.0){\makebox(1,1){\rule[-1ex]{0ex}{3ex}{\normalsize the}}}

\put(48.611,101.0){\makebox(1,1){\rule[-1ex]{0ex}{3ex}{\normalsize\barr{Adj}}}}
   \put(48.611,96.0){\line(0,-1){16.0}}
   \put(48.611,71.0){\makebox(1,1){\rule[-1ex]{0ex}{3ex}{\normalsize Adj}}}
   \put(48.611,66.0){\line(0,-1){46.0}}
   \put(48.611,11.0){\makebox(1,1){\rule[-1ex]{0ex}{3ex}{\normalsize old}}}
   \put(87.5,101.0){\makebox(1,1){\rule[-1ex]{0ex}{3ex}{\normalsize N}}}
   \put(87.5,96.0){\line(0,-1){76.0}}
   \put(87.5,11.0){\makebox(1,1){\rule[-1ex]{0ex}{3ex}{\normalsize dog}}}
   \put(131.25,131.0){\makebox(1,1){\rule[-1ex]{0ex}{3ex}{\normalsize V}}}
   \put(131.25,126.0){\line(0,-1){106.0}}
   \put(131.25,11.0){\makebox(1,1){\rule[-1ex]{0ex}{3ex}{\normalsize chased}}}

\put(194.444,131.0){\makebox(1,1){\rule[-1ex]{0ex}{3ex}{\normalsize\barr{N}}}}
   \put(194.444,126.0){\line(-6,-5){19.444}}
   \put(194.444,126.0){\line(6,-5){19.445}}
   \put(175.0,101.0){\makebox(1,1){\rule[-1ex]{0ex}{3ex}{\normalsize
\barr{D}}}}
   \put(175.0,96.0){\line(0,-1){16.0}}
   \put(175.0,71.0){\makebox(1,1){\rule[-1ex]{0ex}{3ex}{\normalsize D}}}
   \put(175.0,66.0){\line(0,-1){46.0}}
   \put(175.0,11.0){\makebox(1,1){\rule[-1ex]{0ex}{3ex}{\normalsize the}}}
   \put(213.889,101.0){\makebox(1,1){\rule[-1ex]{0ex}{3ex}{\normalsize N}}}
   \put(213.889,96.0){\line(0,-1){76.0}}
   \put(213.889,11.0){\makebox(1,1){\rule[-1ex]{0ex}{3ex}{\normalsize cat}}}
   \put(283.16,131.0){\makebox(1,1){\rule[-1ex]{0ex}{3ex}{\normalsize
\barr{P}}}}
   \put(283.16,126.0){\line(-2,-1){30.382}}
   \put(283.16,126.0){\line(2,-1){30.382}}
   \put(252.778,101.0){\makebox(1,1){\rule[-1ex]{0ex}{3ex}{\normalsize P}}}
   \put(252.778,96.0){\line(0,-1){76.0}}
   \put(252.778,11.0){\makebox(1,1){\rule[-1ex]{0ex}{3ex}{\normalsize into}}}

\put(313.542,101.0){\makebox(1,1){\rule[-1ex]{0ex}{3ex}{\normalsize\barr{N}}}}
   \put(313.542,96.0){\line(-4,-3){21.875}}
   \put(313.542,96.0){\line(4,-3){21.875}}

\put(291.667,71.0){\makebox(1,1){\rule[-1ex]{0ex}{3ex}{\normalsize\barr{D}}}}
   \put(291.667,66.0){\line(0,-1){16.0}}
   \put(291.667,41.0){\makebox(1,1){\rule[-1ex]{0ex}{3ex}{\normalsize D}}}
   \put(291.667,36.0){\line(0,-1){16.0}}
   \put(291.667,11.0){\makebox(1,1){\rule[-1ex]{0ex}{3ex}{\normalsize the}}}
   \put(335.417,71.0){\makebox(1,1){\rule[-1ex]{0ex}{3ex}{\normalsize N}}}
   \put(335.417,66.0){\line(0,-1){46.0}}
   \put(335.417,11.0){\makebox(1,1){\rule[-1ex]{0ex}{3ex}{\normalsize garden}}}
   \put(0.0,0.0){\makebox(1,1)[l]{\rule[-1ex]{0ex}{3ex}{\normalsize}}}
  \end{picture}
\caption{X--bar translation of the structure in Figure 1.}
\end{figure}
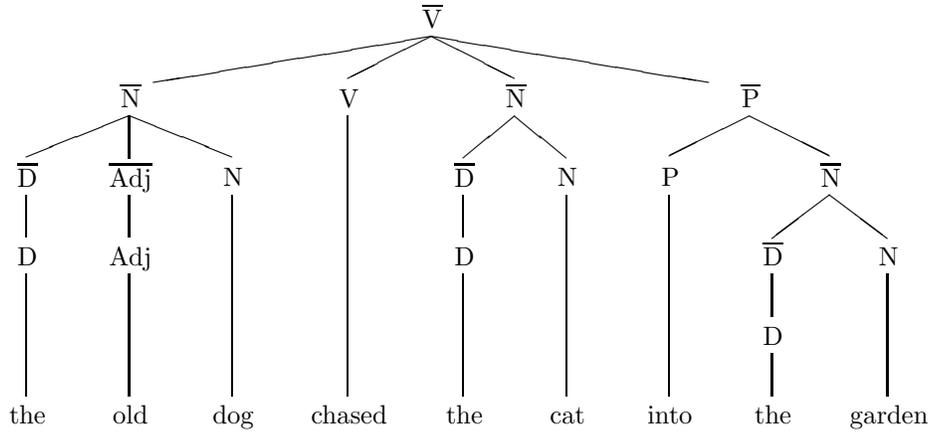

\section{Difficulty 1: The proform {\em one}}

Dependency grammar runs into substantial difficulty trying to account
for the proform {\em one.}  The generalization to be captured is that
{\em one} stands for a constituent larger than the N but smaller than
the NP:
\begin{flushleft} \em
a young long--haired \underline{student} and an older short--haired
\underline{one} \\
a young \underline{long--haired student} and an older \underline{one} \\
a \underline{young long--haired student} and another \underline{one} \\
\end{flushleft}
The standard X--bar analysis (Andrews 1983, Radford 1988:189) accounts for
this behavior elegantly by postulating that {\em one} is a pro-\barr{N},
and that \barr{N}'s form stacked structures (Figure 3).
Dependency grammar can do no such thing, because in dependency grammar
as normally understood, all the modifiers hang from the same \barr{N}
node (Figure 4).

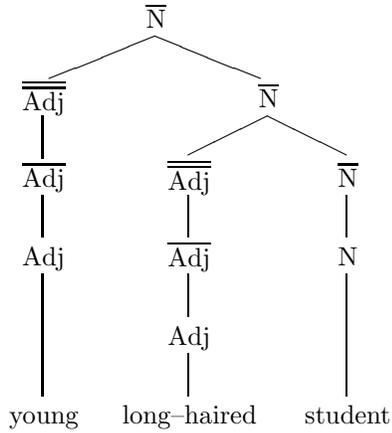
\begin{figure}
  \begin{picture}(350.0,174.0)(0,0)
   \put(62.5,161.0){\makebox(1,1){\rule[-1ex]{0ex}{3ex}{\normalsize \barr{N}}}}
   \put(62.5,156.0){\line(-5,-2){40}}
   \put(62.5,156.0){\line(5,-2){40}}
   \put(20.0,131.0){\makebox(1,1){\rule[-1ex]{0ex}{3ex}{\normalsize
\barrr{Adj}}}}
   \put(20.0,126.0){\line(0,-1){16.0}}
   \put(20.0,101.0){\makebox(1,1){\rule[-1ex]{0ex}{3ex}{\normalsize
\barr{Adj}}}}
   \put(20.0,96.0){\line(0,-1){16.0}}
   \put(20.0,71.0){\makebox(1,1){\rule[-1ex]{0ex}{3ex}{\normalsize Adj}}}
   \put(20.0,66.0){\line(0,-1){46.0}}
   \put(20.0,11.0){\makebox(1,1){\rule[-1ex]{0ex}{3ex}{\normalsize young}}}
   \put(105.0,131.0){\makebox(1,1){\rule[-1ex]{0ex}{3ex}{\normalsize
\barr{N}}}}
   \put(105.0,126.0){\line(-2,-1){30.0}}
   \put(105.0,126.0){\line(2,-1){30.0}}
   \put(75.0,101.0){\makebox(1,1){\rule[-1ex]{0ex}{3ex}{\normalsize
\barrr{Adj}}}}
   \put(75.0,96.0){\line(0,-1){16.0}}
   \put(75.0,71.0){\makebox(1,1){\rule[-1ex]{0ex}{3ex}{\normalsize
\barr{Adj}}}}
   \put(75.0,66.0){\line(0,-1){16.0}}
   \put(75.0,41.0){\makebox(1,1){\rule[-1ex]{0ex}{3ex}{\normalsize Adj}}}
   \put(75.0,36.0){\line(0,-1){16.0}}
   \put(75.0,11.0){\makebox(1,1){\rule[-1ex]{0ex}{3ex}{\normalsize
long--haired}}}
   \put(135.0,101.0){\makebox(1,1){\rule[-1ex]{0ex}{3ex}{\normalsize
\barr{N}}}}
   \put(135.0,96.0){\line(0,-1){16.0}}
   \put(135.0,71.0){\makebox(1,1){\rule[-1ex]{0ex}{3ex}{\normalsize N}}}
   \put(135.0,66.0){\line(0,-1){46.0}}
   \put(135.0,11.0){\makebox(1,1){\rule[-1ex]{0ex}{3ex}{\normalsize student}}}
   \put(0.0,0.0){\makebox(1,1)[l]{\rule[-1ex]{0ex}{3ex}{\normalsize}}}
  \end{picture}
\caption{{\em One} can stand for any of the three \barr{N}'s here.}
\end{figure}

\begin{figure}
\begin{picture}(145,110)
\put(125,50){\line(0,1){60}}
\put(125,40){\makebox(0,0)[b]{N}}
\put(125,35){\line(0,-1){15}}
\put(125,10){\makebox(0,0)[b]{student}}
\put(125,110){\line(-3,-2){60}}
\put(65,50){\line(0,1){20}}
\put(65,40){\makebox(0,0)[b]{Adj}}
\put(65,35){\line(0,-1){15}}
\put(65,10){\makebox(0,0)[b]{long--haired}}
\put(125,110){\line(-3,-1){120}}
\put(5,50){\line(0,1){20}}
\put(5,40){\makebox(0,0)[b]{Adj}}
\put(5,35){\line(0,-1){15}}
\put(5,10){\makebox(0,0)[b]{young}}
\end{picture}
\hfill
\begin{picture}(5.0,50.0)
   \put(2.5,48.0){\makebox(0,0)[b]{\Large $=$}}
\end{picture}
\hfill
 \begin{picture}(145.0,84.0)(0,0)
   \put(70.0,101.0){\makebox(1,1){\rule[-1ex]{0ex}{3ex}{\normalsize \barr{N}}}}
   \put(70.0,96.0){\line(-4,-1){55.0}}
   \put(70.0,96.0){\line(0,-1){16.0}}
   \put(70.0,96.0){\line(4,-1){60.0}}
   \put(15.0,71.0){\makebox(1,1){\rule[-1ex]{0ex}{3ex}{\normalsize
\barr{Adj}}}}
   \put(15.0,66.0){\line(0,-1){16.0}}
   \put(15.0,41.0){\makebox(1,1){\rule[-1ex]{0ex}{3ex}{\normalsize Adj}}}
   \put(15.0,36.0){\line(0,-1){16.0}}
   \put(15.0,11.0){\makebox(1,1){\rule[-1ex]{0ex}{3ex}{\normalsize young}}}
   \put(70.0,71.0){\makebox(1,1){\rule[-1ex]{0ex}{3ex}{\normalsize
\barr{Adj}}}}
   \put(70.0,66.0){\line(0,-1){16.0}}
   \put(70.0,41.0){\makebox(1,1){\rule[-1ex]{0ex}{3ex}{\normalsize Adj}}}
   \put(70.0,36.0){\line(0,-1){16.0}}
   \put(70.0,11.0){\makebox(1,1){\rule[-1ex]{0ex}{3ex}{\normalsize
long--haired}}}
   \put(130.0,71.0){\makebox(1,1){\rule[-1ex]{0ex}{3ex}{\normalsize N}}}
   \put(130.0,66.0){\line(0,-1){46.0}}
   \put(130.0,11.0){\makebox(1,1){\rule[-1ex]{0ex}{3ex}{\normalsize student}}}
   \put(0.0,0.0){\makebox(1,1)[l]{\rule[-1ex]{0ex}{3ex}{\normalsize}}}
  \end{picture}
\caption{This dependency analysis (shown with its X--bar equivalent)
lacks the stacked \barr{N}'s needed to account for {\em one}.}
\end{figure}
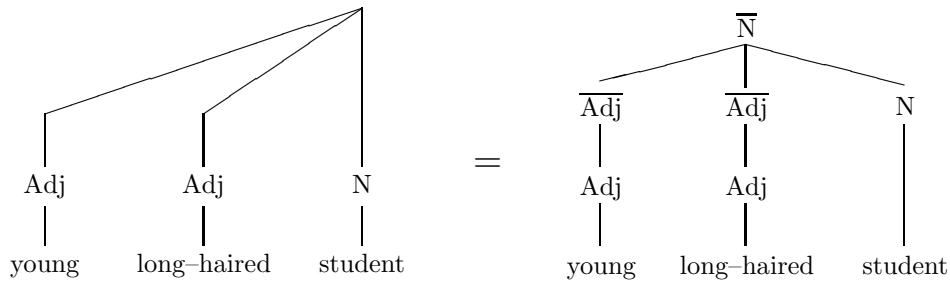

Further, the stacked \barr{N} analysis predicts a structural ambiguity
if there are modifiers on both sides of the head noun --- and the
behavior of {\em one} shows that this ambiguity is real.
Each \barr{N} in either tree in Figure 5 can be the antecedent of {\em
one}:
\begin{flushleft} \em
the long--haired \underline{student} from Cambridge
and a short--haired \underline{one} from Oxford\\
the long--haired \underline{student from Cambridge}
and a short--haired \underline{one}\\
the \underline{long--haired student} from Cambridge
and \underline{one} from Oxford\\
this \underline{long--haired student from Cambridge}
and the other \underline{one}
\end{flushleft}
Again dependency grammar is left high and dry --- DG formalism can
recognize neither the stacking nor the ambiguity, because all the
modifiers have the same head.

\begin{figure}
\begin{flushleft} \xpt
  \begin{picture}(350.0,174.0)(0,0)
   \put(77.5,161.0){\makebox(1,1){\rule[-1ex]{0ex}{3ex}{\xpt
$\overline{\overline{\rm \xpt N}}$}}}
   \put(77.5,156.0){\line(-4,-1){67.5}}
   \put(77.5,156.0){\line(4,-1){67.5}}
   \put(10.0,131.0){\makebox(1,1){\rule[-1ex]{0ex}{3ex}{\xpt D}}}
   \put(10.0,126.0){\line(0,-1){106.0}}
   \put(10.0,11.0){\makebox(1,1){\rule[-1ex]{0ex}{3ex}{\xpt the}}}
   \put(145.0,131.0){\makebox(1,1){\rule[-1ex]{0ex}{3ex}{\xpt $\overline{\rm
\xpt N}$}}}
   \put(145.0,126.0){\line(-3,-1){47.0}}
   \put(145.0,126.0){\line(3,-1){47.0}}
   \put(95.0,101.0){\makebox(1,1){\rule[-1ex]{0ex}{3ex}{\xpt $\overline{\rm
\xpt N}$}}}
   \put(95.0,96.0){\line(-2,-1){30.0}}
   \put(95.0,96.0){\line(2,-1){30.0}}
   \put(65.0,71.0){\makebox(1,1){\rule[-1ex]{0ex}{3ex}{\xpt \barrr{Adj}}}}
   \put(65.0,66.0){\line(-1,-2){23.75}}
   \put(65.0,66.0){\line(1,-2){23.75}}
   \put(41.25,18.5){\line(1,0){47.5}}
   \put(65.0,11.0){\makebox(1,1){\rule[-1ex]{0ex}{3ex}{\xpt long--haired}}}
   \put(125.0,71.0){\makebox(1,1){\rule[-1ex]{0ex}{3ex}{\xpt $\overline{\rm
\xpt N}$}}}
   \put(125.0,66.0){\line(0,-1){16.0}}
   \put(125.0,41.0){\makebox(1,1){\rule[-1ex]{0ex}{3ex}{\xpt N}}}
   \put(125.0,36.0){\line(0,-1){16.0}}
   \put(125.0,11.0){\makebox(1,1){\rule[-1ex]{0ex}{3ex}{\xpt student}}}
   \put(195.0,101.0){\makebox(1,1){\rule[-1ex]{0ex}{3ex}{\xpt \barrr{P}}}}
   \put(195.0,96.0){\line(-2,-5){31.0}}
   \put(195.0,96.0){\line(2,-5){31.0}}
   \put(164.0,18.5){\line(1,0){62.0}}
   \put(195.0,11.0){\makebox(1,1){\rule[-1ex]{0ex}{3ex}{\xpt from Cambridge}}}
   \put(0.0,0.0){\makebox(1,1)[l]{\rule[-1ex]{0ex}{3ex}{\xpt}}}
  \end{picture}
{}~\\
{}~\\
  \begin{picture}(350.0,174.0)(0,0)
   \put(61.25,161.0){\makebox(1,1){\rule[-1ex]{0ex}{3ex}{\xpt
$\overline{\overline{\rm \xpt N}}$}}}
   \put(61.25,156.0){\line(-3,-1){51.25}}
   \put(61.25,156.0){\line(3,-1){51.25}}
   \put(10.0,131.0){\makebox(1,1){\rule[-1ex]{0ex}{3ex}{\xpt D}}}
   \put(10.0,126.0){\line(0,-1){106.0}}
   \put(10.0,11.0){\makebox(1,1){\rule[-1ex]{0ex}{3ex}{\xpt the}}}
   \put(112.5,131.0){\makebox(1,1){\rule[-1ex]{0ex}{3ex}{\xpt $\overline{\rm
\xpt N}$}}}
   \put(112.5,126.0){\line(-3,-1){44.5}}
   \put(112.5,126.0){\line(3,-1){44.5}}
   \put(65.0,101.0){\makebox(1,1){\rule[-1ex]{0ex}{3ex}{\xpt \barrr{Adj}}}}
   \put(65.0,96.0){\line(-1,-3){24.75}}
   \put(65.0,96.0){\line(1,-3){24.75}}
   \put(40.25,21.75){\line(1,0){49.5}}
   \put(65.0,11.0){\makebox(1,1){\rule[-1ex]{0ex}{3ex}{\xpt long--haired}}}
   \put(160.0,101.0){\makebox(1,1){\rule[-1ex]{0ex}{3ex}{\xpt $\overline{\rm
\xpt N}$}}}
   \put(160.0,96.0){\line(-2,-1){32.0}}
   \put(160.0,96.0){\line(2,-1){32.0}}
   \put(125.0,71.0){\makebox(1,1){\rule[-1ex]{0ex}{3ex}{\xpt $\overline{\rm
\xpt N}$}}}
   \put(125.0,66.0){\line(0,-1){16.0}}
   \put(125.0,41.0){\makebox(1,1){\rule[-1ex]{0ex}{3ex}{\xpt N}}}
   \put(125.0,36.0){\line(0,-1){16.0}}
   \put(125.0,11.0){\makebox(1,1){\rule[-1ex]{0ex}{3ex}{\xpt student}}}
   \put(195.0,71.0){\makebox(1,1){\rule[-1ex]{0ex}{3ex}{\xpt \barrr{P}}}}
   \put(195.0,66.0){\line(-2,-3){31.5}}
   \put(195.0,66.0){\line(2,-3){31.5}}
   \put(163.5,18.75){\line(1,0){63.0}}
   \put(195.0,11.0){\makebox(1,1){\rule[-1ex]{0ex}{3ex}{\xpt from Cambridge}}}
   \put(0.0,0.0){\makebox(1,1)[l]{\rule[-1ex]{0ex}{3ex}{\xpt}}}
  \end{picture}
{}~\\
{}~\\
\end{flushleft}
\caption{Dependency grammar cannot express this structural ambiguity; DG
can only say that both {\em long--haired} and {\em from Cambridge}
modify {\em student}.}
\end{figure}

\section{Difficulty 2: Semantics of multiple modifiers}

A second difficulty with dependency grammar comes from semantics.
Dahl (1980) points out that proximity to the head affects the meaning of
certain modifiers.  A {\em typical French house} is something typical of
French houses, not merely a house that is French and typical.
Semantically, at least, its structure is therefore:
\begin{flushleft} \em
$[$ typical $[$ French house $]$ $]$
\end{flushleft}
which is consistent with a stacked \barr{N} analysis.  But this grouping
cannot be expressed by dependency grammar, because as far as DG is
concerned, {\em typical} and {\em French} are dependents of {\em house},
and there is no intermediate syntactic structure.

Andrews (1983) points out that the same thing happens with verbs.
Contrast:
\begin{flushleft}
{\em $[$ $[$ knocked twice $]$ intentionally $]$}
\hfill (acted on one intention, to knock twice)\\
{\em $[$ $[$ knocked intentionally $]$ twice $]$}
\hfill (had the intention two times)
\end{flushleft}
These argue strongly for stacking of \barr{V}'s, or at least for
something comparable on the semantic level.

Note by the way that if
there are modifiers on both sides of the verb, an ambiguity arises just
as it did with nouns: {\em intentionally knocked twice} is ambiguous
between {\em $[$ $[$ intentionally knocked $]$ twice $]$} and
{\em $[$ intentionally $[$ knocked twice $]$ $]$}.

Crucially, these phenomena entail that if one adopts a non--stacked
syntax such as that mandated by the standard interpretation of DG, then
the semantic component of the grammar must know not only the grammatical
relations recognized by the syntax, but also the comparative proximity
of the various modifiers to the head.

\section{Reinterpreting dependency grammar}

Dependency grammar can be salvaged from this mess by reinterpreting its
claims about phrase structure.  Recall that in a dependency grammar,
constituency is a defined concept.  The solution is therefore to change
the definition.  Specifically, instead of being considered equivalent to
flat X--bar trees, dependency structures can be mapped onto X--bar trees
that introduce stacking in a principled way.%
\footnote{This reinterpretation was suggested by Hudson's proposal
(1980b:499--501, 1990:149--150)
that the semantic effect of proximity of the head is due to a parsing
effect. Since parsing is nothing if not syntactic, it seems desirable to
incorporate this proposal into the syntactic theory.}

Here is a sketch of such a reinterpretation, consistent with current
X--bar theory.  Given a head (X) and its dependents, attach the
dependents to the head by forming stacked \barr{X} nodes as follows:
\begin{enumerate}
\item Attach subcategorized complements first, all under the same
\barr{X} node.%
\footnote{Actually, it is immaterial to my argument whether all the
complements hang from the same node or whether they, too, are introduced
by binary branching, like the adjuncts.}
If there are none, create the \barr{X} node anyway.
\item Then attach modifiers, one at a time, by working outward from the
one nearest the head noun, and adding a stacked \barr{X} node for each.
\item Finally, create an \barrr{X} node at the top of the stack, and
attach the specifier (determiner), if any.
\end{enumerate}
Thus the dependency structure
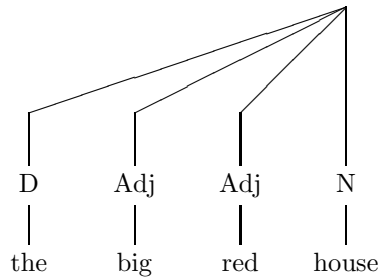
\begin{flushleft}
\begin{picture}(145,110)
\put(125,50){\line(0,1){60}}
\put(125,40){\makebox(0,0)[b]{N}}
\put(125,35){\line(0,-1){15}}
\put(125,10){\makebox(0,0)[b]{house}}
\put(125,110){\line(-1,-1){40}}
\put(85,50){\line(0,1){20}}
\put(85,40){\makebox(0,0)[b]{Adj}}
\put(85,35){\line(0,-1){15}}
\put(85,10){\makebox(0,0)[b]{red}}
\put(125,110){\line(-2,-1){80}}
\put(45,50){\line(0,1){20}}
\put(45,40){\makebox(0,0)[b]{Adj}}
\put(45,35){\line(0,-1){15}}
\put(45,10){\makebox(0,0)[b]{big}}
\put(125,110){\line(-3,-1){120}}
\put(5,50){\line(0,1){20}}
\put(5,40){\makebox(0,0)[b]{D}}
\put(5,35){\line(0,-1){15}}
\put(5,10){\makebox(0,0)[b]{the}}
\end{picture}
\end{flushleft}
maps, under the new interpretation, to the stacked structure:
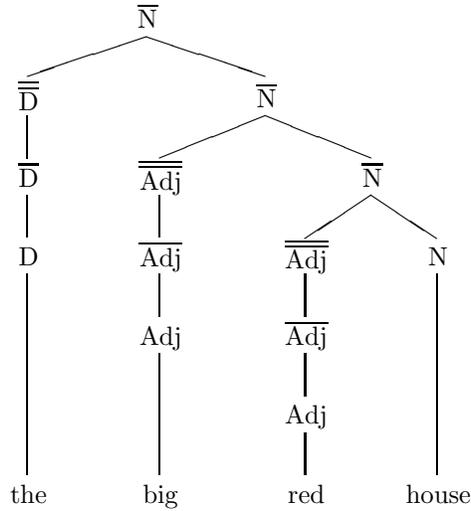
\begin{flushleft}
  \begin{picture}(350.0,204.0)(0,0)
   \put(60.0,191.0){\makebox(1,1){\rule[-1ex]{0ex}{3ex}{\normalsize \barr{N}}}}
   \put(60.0,186.0){\line(-3,-1){45.0}}
   \put(60.0,186.0){\line(3,-1){45.0}}
   \put(15.0,161.0){\makebox(1,1){\rule[-1ex]{0ex}{3ex}{\normalsize
\barrr{D}}}}
   \put(15.0,156.0){\line(0,-1){16.0}}
   \put(15.0,131.0){\makebox(1,1){\rule[-1ex]{0ex}{3ex}{\normalsize \barr{D}}}}
   \put(15.0,126.0){\line(0,-1){16.0}}
   \put(15.0,101.0){\makebox(1,1){\rule[-1ex]{0ex}{3ex}{\normalsize D}}}
   \put(15.0,96.0){\line(0,-1){76.0}}
   \put(15.0,11.0){\makebox(1,1){\rule[-1ex]{0ex}{3ex}{\normalsize the}}}
   \put(105.0,161.0){\makebox(1,1){\rule[-1ex]{0ex}{3ex}{\normalsize
\barr{N}}}}
   \put(105.0,156.0){\line(-5,-2){40.0}}
   \put(105.0,156.0){\line(5,-2){40.0}}
   \put(65.0,131.0){\makebox(1,1){\rule[-1ex]{0ex}{3ex}{\normalsize
\barrr{Adj}}}}
   \put(65.0,126.0){\line(0,-1){16.0}}
   \put(65.0,101.0){\makebox(1,1){\rule[-1ex]{0ex}{3ex}{\normalsize
\barr{Adj}}}}
   \put(65.0,96.0){\line(0,-1){16.0}}
   \put(65.0,71.0){\makebox(1,1){\rule[-1ex]{0ex}{3ex}{\normalsize Adj}}}
   \put(65.0,66.0){\line(0,-1){46.0}}
   \put(65.0,11.0){\makebox(1,1){\rule[-1ex]{0ex}{3ex}{\normalsize big}}}
   \put(145.0,131.0){\makebox(1,1){\rule[-1ex]{0ex}{3ex}{\normalsize
\barr{N}}}}
   \put(145.0,126.0){\line(-3,-2){25.0}}
   \put(145.0,126.0){\line(3,-2){25.0}}
   \put(120.0,101.0){\makebox(1,1){\rule[-1ex]{0ex}{3ex}{\normalsize
\barrr{Adj}}}}
   \put(120.0,96.0){\line(0,-1){16.0}}
   \put(120.0,71.0){\makebox(1,1){\rule[-1ex]{0ex}{3ex}{\normalsize
\barr{Adj}}}}
   \put(120.0,66.0){\line(0,-1){16.0}}
   \put(120.0,41.0){\makebox(1,1){\rule[-1ex]{0ex}{3ex}{\normalsize Adj}}}
   \put(120.0,36.0){\line(0,-1){16.0}}
   \put(120.0,11.0){\makebox(1,1){\rule[-1ex]{0ex}{3ex}{\normalsize red}}}
   \put(170.0,101.0){\makebox(1,1){\rule[-1ex]{0ex}{3ex}{\normalsize N}}}
   \put(170.0,96.0){\line(0,-1){76.0}}
   \put(170.0,11.0){\makebox(1,1){\rule[-1ex]{0ex}{3ex}{\normalsize house}}}
   \put(0.0,0.0){\makebox(1,1)[l]{\rule[-1ex]{0ex}{3ex}{\normalsize}}}
  \end{picture}
\end{flushleft}
The distinction between specifier, modifier, and complement is already
needed in dependency grammar, so this interpretation does not require
anything new in the dependency formalism (Hudson 1990:202--211).

Note that if there are modifiers both before and after the head, the
resulting X--bar tree is not unique --- and this non--uniqueness is
desirable, because the resulting alternatives, such as
\begin{flushleft} \em \hsize=7.5in
$[$ $[$ long--haired student $]$ from Cambridge $]$ $:$ $[$ long--haired
$[$ student from Cambridge $]$ $]$\\
{}~\\
$[$ $[$ intentionally knocked $]$ twice $]$ $:$ $[$ intentionally $[$ knocked
twice $]$ $]$
\end{flushleft}
are exactly the ones required by the evidence.

\section{Conclusion}

The alert reader may wonder, at this point, whether dependency grammar
has been salvaged or rather refuted, because under the new
interpretation, DG is a notational variant of current X--bar theory.
To this I have several replies:
\begin{enumerate}
\item It should not be surprising when separate theories of the same
phenomena develop convergently.
\item DG always {\sc was} a notational variant of X--bar theory; I have
merely brought its implicit X--bar theory up to date.
\item DG still imposes stricter requirements than transformational
grammar, because in DG, violations of X--bar theory are flatly
impossible, not just undesirable.
\end{enumerate}
In any case, the dependency perspective on sentence structure has proved
its worth not only in syntactic theorizing, but also in language
teaching, parsing, and other practical applications.  Indeed, dependency
concepts, such as government and c--command, are becoming increasingly
prominent in transformational grammar.  Dependency grammar
can complement other approaches to syntax in much the same way that
relational grammar, fifteen years ago, provided an organizing
perspective on what had previously been a heterogeneous set of
syntactic transformations.

\section*{References}
\begin{reflist}
Andrews, Avery, III (1983) A note on the constituent structure of
modifiers. {\em Linguistic Inquiry} 14:695--697.

Covington, Michael A. (1990) Parsing discontinuous constituents in
dependency grammar. {\em Computational Linguistics} 16:234--236.

Dahl, \"{O}sten (1980) Some arguments for higher nodes in syntax: a
reply to Hudson's `Constituency and dependency.' {\em Linguistics}
18:485--488.

Hudson, Richard (1980a) Constituency and dependency. {\em Linguistics}
18:179--198.

Hudson, Richard (1980b) A second attack on constituency: a reply to
Dahl.  {\em Linguistics} 18:489--504.

Hudson, Richard (1990) {\em English word grammar.}
Oxford: Basil Blackwell.

Mel'\v{c}uk, Igor A. (1987) {\em Dependency syntax: theory and practice.}
Albany: State University of New York Press.

Radford, Andrew (1988) {\em Transformational grammar.}  Cambridge:
Cambridge University Press.

Robinson, Jane J. (1970) Dependency structures and transformational
rules. {\em Language} 46:259--285.

Starosta, Stanley (1988) {\em The case for lexicase.}
London: Pinter.

Tesni\`{e}re, Lucien (1959) {\em \'{E}l\'{e}ments de syntaxe structurale.}
Paris: Klincksieck.

\end{reflist}
\end{document}